\pdfoutput=1
\documentclass[11pt]{article}
\usepackage[utf8]{inputenc}
\usepackage{xcolor} % For color customization
\usepackage{tcolorbox} % For the colored box
\tcbset{
    keytakeaways/.style={
        colback=blue!1, % Background color
        colframe=blue!30, % Frame color
        fonttitle=\bfseries, % Title font style
        title=Key Takeaways, % Default title
        boxrule=0.5mm, % Thickness of the frame
        arc=4mm, % Rounded corners
        left=1mm, % Left padding
        right=1mm, % Right padding
        top=1mm, % Top padding
        bottom=1mm, % Bottom padding
        coltitle=black % Title text color
    }
}

\usepackage[]{acl}

\usepackage{times}
\usepackage{multicol}

\usepackage{amsmath}
\usepackage{array}
\usepackage{latexsym}
\usepackage{enumitem}
\usepackage{graphicx}
\usepackage{fontawesome5}
\usepackage{arydshln}

\definecolor{humanColor}{HTML}{A8E6CF} % Light teal for Humans
\definecolor{plmColor}{HTML}{D4C5F9}   % Light purple for PLMs
\definecolor{evalColor}{HTML}{D1E9FC}  % Light blue for Evaluation
\usepackage[T1]{fontenc}
\usepackage[utf8]{inputenc}

\usepackage{microtype}

\usepackage{inconsolata}

\usepackage{graphicx}

\title{The potential -- and the pitfalls -- of using pre-trained language models as cognitive science theories}

\newcommand{\gtlogo}{\raisebox{3.4pt}{\includegraphics[scale=0.04]{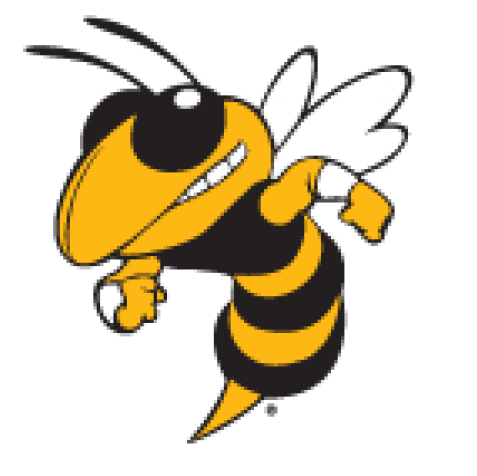}}}

\author{Raj Sanjay Shah, Sashank Varma
 \\ 
 Georgia Institute of Technology \gtlogo \\
 \textcolor{darkblue}{{\{\href{mailto:rajsanjayshah@gatech.edu}{rajsanjayshah},\href{mailto:varma@gatech.edu}{varma}\}@gatech.edu}
}}

\begin{document}
\maketitle
\begin{abstract}
Many studies
%%% RESOLVED %%% SV: Commented out the next bit of text to compact the first and second sentences into one. They were redundant, and in fact defined the PLM acronym twice!
% show evidence for cognitive abilities in Pre-trained Language Models (PLMs). Researchers 
have evaluated the cognitive alignment of Pre-trained Language Models (PLMs), i.e., their correspondence to adult performance across a range of cognitive domains. Recently, the focus has expanded to the developmental alignment of these models: identifying phases during training where improvements in model performance track improvements in children’s thinking over development. 
% \raj{I propose we replace the following two points with the next text below:}
% However, the challenges to the use of PLMs as cognitive science theories are twofold: (1) PLMs have very different architectures than human minds and brains, and the data sets on which they are trained differ in many ways from the inputs children receive. (2) The “outputs” of PLMs are different from the behavioral measures that cognitive scientists collect in their experiments and evaluate their theories against. 
%%% SV: The next sentence is new text you wrote. I like it so I have fully included it.
However, there are many challenges to the use of PLMs as cognitive science theories, including different architectures, different training data modalities and scales, and limited model interpretability.
In this paper, we distill lessons learned from treating PLMs, not as engineering artifacts but as cognitive science and developmental science models. We review assumptions used by researchers to map measures of PLM performance to measures of human performance.
%%% RESOLVED SV: Not sure the italics are needed in the final sentence.
We identify potential pitfalls of this approach to understanding human thinking, and we end by enumerating criteria for using PLMs as credible accounts of cognition and cognitive development. 
\end{abstract}

% \sv{Another, harkening back to a famous paper on the "seductive allure of neuroscience explanations" in thinking about thinking: The seductive allure of language models: Lessons from using PLMs as cognitive science theories}

% \raj{I would prefer mine or Pre-trained Language models are harsh mistresses: The potential (and the pitfalls) of using PLMs as cognitive science theories or the reference to the famous paper. Alternatively, we can keep the lame title for the reviewing cycle and change it up later - "The potential (and the pitfalls) of using PLMs as cognitive science theories"}

% \sv{In my fields, I would be slightly worried that words like "mistress" and "seductive" might make female reviewers mad, and omit them. My rank order: (1) The potential -- and the pitfalls -- of using PLMs as cognitive science theories. (2) Pre-trained Language models are harsh mistresses: The potential (and the pitfalls) of using PLMs as cognitive science theories. (3) Language Models are a cognitive scientist's seductive mistress: Lessons from using PLMs for Human Cognitive Modeling. Let's not use BOTH "mistress(es)" and "seductive"!}

\section{Introduction}

With the improving performance of pre-trained language models \cite{touvron2023llama,geminiteam2023gemini,openai2023gpt4, wei2022emergent}, researchers are increasingly advocating for their use as computational models of cognition \cite{piantadosi2023modern,mahowald2024dissociating,warstadt2024artificial, coda2024cogbench}.
This is true for many domains including mathematical reasoning \cite{shah2023numeric, ahn2024large}, language comprehension \cite{warstadt2020blimp, cog_sci_garden_path,
% ye2023comprehensive, koubaa2023gpt,
hu2024language}, concept understanding \cite{cog_sci_typicallity}, spatial reasoning \cite{ramakrishnan2024does} and analogical reasoning \cite{webb2023emergent, hu2023context}. For example, \citet{shah2023numeric} investigated the latent number representations of PLMs, finding that they showed the \emph{distance}, \emph{size}, and \emph{ratio} effects observed in humans and understood to be the behavioral signatures of a ``mental number line'' \cite{moyerTimeRequiredJudgements1967,parkman1971temporal, halberda2008individual}.
% \raj{will change this example to something (self note)} 
% \sv{Two more cites: (1) Parkman, J. M. (1971). Temporal aspects of digit and letter inequality judgments. Journal of Experimental Psychology, 91, 191-205. (2) Halberda, J., Mazzocco, M. M., \& Feigenson, L. (2008). Individual differences in non-verbal number acuity correlate with maths achievement. Nature, 455, 665-668.} 
To take another example, Raven's Progressive Matrices test is the standard psychometric measure of \emph{fluid reasoning}. 
Although Raven's problems are visual, \citet{webb2023emergent} translated
% \raj{should we use the word "adapted"} 
them to equivalent "digit encodings" and showed that PLMs perform as well as humans on this test. More generally, many works combine tests across multiple task domains to form comprehensive benchmarks that enable scientific evaluation of the alignment of ML models to human cognition \cite{chang2024benchmarking, zhuang2023efficiently,shah2024development,coda2024cogbench, wang2024coglm,tan2024devbench}.
% \sv{I can pick out a few examples and present their findings, as you do in the next paragraph for developmental studies.}\raj{sounds good} \sv{OK, I made these edits, adding your 2023 paper on number and Webb's 2023 paper on Ravens (since it is a nice example of 'adaptation' the problem presentation.}\raj{Thank you so much}

Recently, researchers have begun using PLMs to additionally model the development of cognition in children \cite{hosseini2022artificial, kosoy2023comparing,frank2023bridging, shah2024development, tan2024devbench}. For example, \citet{Portelance2023PredictingAO} suggest the use of language models to predict the age of acquisition of words in children. \citet{wang2024coglm} use Piaget’s Theory of Cognitive Development to estimate that models like GPT-4o show cognitive abilities similar to 20-year-old humans. Instead of just looking at model end states, \citet{shah2024development} investigated developmental trajectories and path dependence: whether the performance improvements of PLMs over training track the growth of cognitive abilities in children over development. Similarly, \citet{tan2024devbench} explored developmental parallels by comparing the learning trajectories of vision-language models to both child and adult behavioral data. To take a final example, researchers have begun varying the exposure of PLMs to multiple languages during pre-training to understand the differential rates of bilingual language development \cite{evanson2023language, marian2023studying, sharma1monolingual}.

\begin{table*}[!h]
    \centering
    \renewcommand{\arraystretch}{1.3}
    \setlength{\tabcolsep}{12pt}
    \resizebox{0.98\textwidth}{!}{%
    \begin{tabular}{>{\arraybackslash}m{0.25\textwidth}:m{0.7\textwidth}}
    \hline

    \multicolumn{2}{c}{\textbf{\large Our Contributions} \faTrophy} \\
    \hline

    \textbf{\faChild ~ Developmental Alignment} & We highlight the alignment of PLMs with human cognitive development and discuss how model checkpoints and training trajectories can mimic children's developmental trajectories. \\
    \hdashline
    \textbf{\faExclamationTriangle ~ Pitfalls of Commission and Omission} & We introduce a new categorization of pitfalls into those of \textit{commission} (methodological mistakes, such as distal linking hypotheses) and those of \textit{omission} (neglecting broader contexts like psychometric and developmental data), providing a newer framework for identifying error types. \\
    \hdashline
    \textbf{\faLink ~ Linking Hypotheses and Measures} & We provide a detailed critique of linking hypotheses, focusing on the need for a robust mapping between model outputs and human cognitive measures, and we warn about distal links that can obscure interpretability and scientific validity. \\
    \hdashline
    
    \textbf{\faMicroscope ~ Promises of Adjacent Fields} & We draw interdisciplinary parallels, such as comparing functional mapping in neuroscience to mechanistic interpretability in PLMs, and discuss how these analogies can generate hypotheses about neural activity. \\
    \hdashline

    \textbf{\faCheck ~ Best Practices for Credibility in Cognitive Science} & We enumerate explicit criteria for using PLMs as credible theories in cognitive science, such as focusing on functional alignment over explanatory adequacy and ensuring sufficiency in mapping between PLMs and human behavior. \\
    \hline
    \end{tabular}
    }
    \caption{Summary of our contributions - distinctions between our work and previous work by \citet{ivanova2023running} and \citet{mahowald2024dissociating}.}
    \label{tab:our_contribution}
\end{table*}

\subsection*{Reader's Guide to the paper}
% \raj{Added emphasis in this paragraph}
In this paper, we describe the pitfalls and promise of using PLMs and candidate theories in cognitive and developmental science. The contributions of this paper are threefold. (1) We first review the \emph{pitfalls} of using PLMs in psychological science and caution researchers against over-interpreting the alignment of PLMS to human cognition and its trajectory over development. (2) Next, we review the \emph{standard assumptions} researchers use to map measures of PLM performance to human performance measures.
%%% RESOLVED SV: As in the Abstract, I am unsure whether we need intalics in the final sentence of this paragraph. Raj: removed the italics
(3) Finally, we build upon previous work \cite{ivanova2023running, mahowald2024dissociating,cuskley2024limitations,birhane2024large} in enumerating best practices for cognitive evaluations of PLMs (refer to table \ref{tab:our_contribution}).
% \cite{zhuang2023efficiently, pellert2023ai, heineman2023rethinking, gennetian2022open} and move towards using PLMs as \emph{credible accounts} of cognitive and developmental science theories.

\section{PLMs as theories in cognitive and developmental science}

% \raj{To add citations here}
\begin{figure*}[h]
\centering
\includegraphics[trim={0cm 0cm 0cm 0cm}, width=0.71\textwidth]{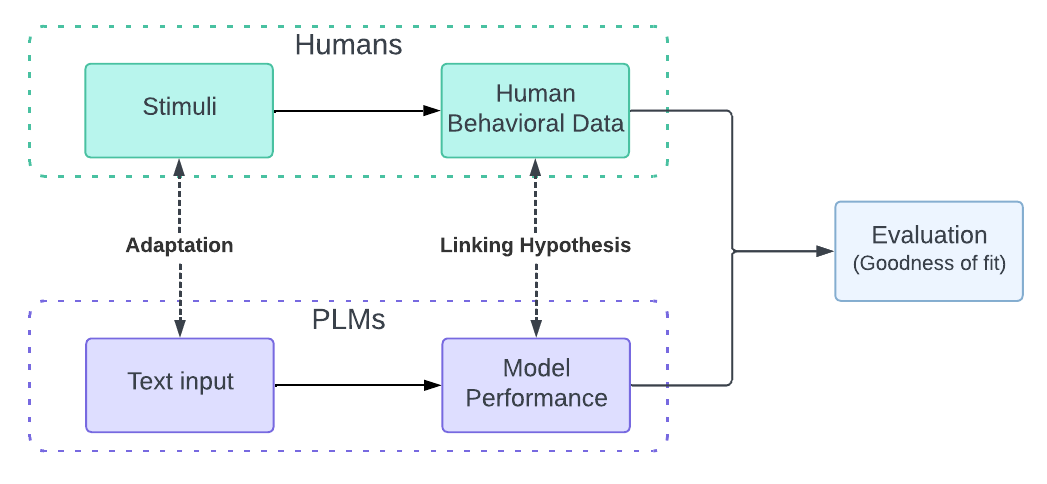}
\caption{The three-stage mapping between human data and model performance when establishing the sufficiency and utility of PLMs as cognitive and developmental science models.}
\label{fig:sufficieency}
\end{figure*}

PLMs make promising candidates as theories of cognitive science due to the nature of language model training \cite{hardy2023large}. PLMs acquire a wide range of capabilities in the pre-training phase, essentially obtaining human-like behaviors “for free” without the need for extensive task-specific tuning or adaptations \cite{demszky2023using,weng2024navigating, yang2024unveiling,minaee2024large,openai2023gpt4}. 
For example, the probabilistic nature of PLM text generation lures researchers to draw parallels with the human decision-making process, where humans anticipate future courses of action based on their past experiences and context. These models can provide insights into how humans may handle ambiguous situations where multiple interpretations are possible \cite{mcgrath2020does,gawlikowski2023survey,dong2024can,belem2024perceptions}.  Moving from cognitive science to developmental science, PLMs have been shown to gradually learn linguistic sensitivities to syntax, semantics, morphology, etc., from mere exposure to (i.e., masked word prediction of) large amounts of text data. This raises the possibility that their learning trajectories mimic the development trajectories of children acquiring language \cite{aher2023using,duan2024hlb,shah2024development}.

This allure of PLMs as theories of cognitive science also extends to adjacent research areas. One example is the ability to simulate personas using PLMs, which enables the proxying of human subjects \cite{park2022social,samuel2024personagym,schuller2024generating,tseng2024two}. This is especially important when real-world experiments with human participants are complex, resource intensive, and/or pose ethical challenges \cite{aher2023using,dillion2023can,hamalainen2023evaluating}. Another adjacent area is neuroscience: Functional mapping in the neuroscience literature (i.e., mapping cognitive functions to their neural correlates) is similar to mechanistic interpretability in PLMs: both map functionalities to structure, provide a component-level analysis, and explain network interactions. This raises the question of whether the functional units of a PLM can be mapped to the functional units of the brain; this would enable their use in generating hypotheses about neural activity \cite{bzdok2024data}.

In its essence, we present an argument for the sufficiency and the utility of PLMs as cognitive and developmental science models: as long as \emph{their performance profiles match those of humans}, then PLMs can be used to \emph{predict} human behavior to specific stimuli. That is, we argue that matching the functional forms of the two is key, and that prediction is the more important criterion than explanation when applying ML models to cognitive phenomena \cite{breiman2003statistical,Yarkoni2017ChoosingPO}.
% \sv{Yarkoni, T., & Westfall, J. (2017). Choosing prediction over explanation in psychology: Lessons from machine learning. Perspectives on Psychological Science,  12, 1100-1122.} \sv{Breiman, L. (2001). Statistical modeling: The two cultures. Statistical Science, 16, 199-231.} 
Prior research in many domains has adopted this sufficiency argument \cite{niu2024large,ong2024gpt}.

% see Table XXX for a comprehensive list of domains and problem spaces where PLMs have been used as cognitive science models. \sv{Let's make this table huge, both to show the sufficiency and soak up the page count. And to increase the value of the paper as a review for guiding those new to the area.}\raj{building in appendix}
 
In human studies, experiments are carefully designed to build and test hypotheses on the effect of cognitive phenomena. Independent of the phenomena under consideration, there are broadly three stages when mapping between the human data and model performance:
\begin{enumerate}
    \item Adaptation of the experimental stimuli and task: The stimuli presented to humans and the task they are asked to perform are modified to match the input modality of PLMs. 
    \item Linking Hypothesis: The model outputs are mapped to indices of human performance. 
    \item Comparison to human performance: Model performance is compared against human performance to evaluate human alignment. This usually involves computing a measure of the \emph{goodness of fit}. 
\end{enumerate}
These three stages are depicted in Figure \ref{fig:sufficieency}). They provide a structure for evaluating the sufficiency or utility of PLMs as cognitive or developmental science theories.

\begin{tcolorbox}[keytakeaways,title=Sufficiency criterion]
   \emph{Under reasonable assumptions} about the three stages -- (1) the adaptation of human stimuli to model inputs, (2) linking hypothesis for the model outputs, and (2) the comparison of model and human performance -- \emph{we are justified in using PLMs to predict human behavior}.
\end{tcolorbox}

%%% SV: This paragraph says a lot of stuff in one long sentence. I would improve it since we revisit generalizability, transferability, and discovery later in the paper. At least divide this sentence into several shorter and more digestible ones. Or else I would comment this text out here and that text there.
% \newtext{This sufficiency criterion extends into two key ideas that promise utility: (1) generalizability and transferability \cite{binz2024centaur}, where demonstrating predictive utility on $n$ tasks builds confidence in performance on the $n+1\text{th}$ task, and (2) discovery, akin to drug discovery research \cite{vattikuti2024improving}, where predictive utility in $n$ tasks generates confident \emph{novel} candidate theories for human cognition based on the model's behavior with novel stimuli.}

This sufficiency criterion extends into two key ideas that offer significant utility. The first is \emph{generalizability and transferability}, as discussed by \citet{binz2024centaur}, demonstrating predictive utility across  $n$ tasks builds confidence in the model’s performance on the 
$n+1\text{th}$ task. The second idea is \emph{discovery}, which parallels drug discovery research, as highlighted by \citet{vattikuti2024improving}. Here, achieving predictive utility across $n$ tasks enables the generation of confident and novel candidate theories for human cognition based on the model's behavior with novel stimuli.
% \newtext{Our sufficiency criteria extends the arguments put forth by \citet{Yarkoni2017ChoosingPO}, advocating that researchers should leverage large datasets and advanced machine learning techniques first to develop models that are sufficient — capable of matching human performance. Only after achieving \emph{sufficiency} should efforts focus on exploring the underlying mechanisms, aiming to distill a parsimonious and interpretable set of principles.}
% Thank you for pointing out the need to expand and clarify our argument here. In greater detail: Over the past two decades, there has been a shift in science from hypothesis-driven to data-driven approaches to model development (Breiman, 2001). Yarkoni and Westfall (2017) made this point forcefully to cognitive scientists, arguing that they should capitalize on large datasets and powerful ML techniques to learn models that are sufficient – that match human performance – first. Only then should they explore why, i.e., distill out a parsimonious set of mechanisms. We agree that this is the best way forward at the current time, with the capabilities of PLMs growing rapidly. Our paper focuses entirely on the first step, establishing that the performance profiles of PLMs match those of humans. This is the three-stage mapping previewed in this section and depicted in Figure 2.

\section{Pitfalls of using PLMs as scientific theories}

% Before articulating the three stages further 
% \raj{ I believe we do not articulate the three stages later, only the linking hypothesis?}, 
Although they show great promise, the assumptions made in the three-stage mapping process can lead to potential pitfalls when using PLMs for cognitive modeling. We draw on current research to enumerate these pitfalls.
%%% SV: The next few sentences are new text you wrote. I like it so I have fully included it.
Many have been noted before in the literature; a new conceptual contribution is to partition them into two distinct classes, pitfalls of commission and pitfalls of omission. Pitfalls of the commission are the methodological and meta-theoretical mistakes researchers may make when comparing PLMs and humans: using relatively distal linking hypotheses, insufficient consideration of the psychological plausibility of training corpora, etc. Pitfalls of omission stem from incorrectly assuming that cognitive ability (e.g., analogical reasoning) is a “module” and failing to consider the high-level context in which it functions: its relation to other cognitive abilities (psychometric data), its progression over-development (cognitive development data), and its neural correlates (neuroscience data).

%%% SV: This text was orphaned. I have commented it out.
%
% First, we consider PLMs as cognitive science theories, i.e., as models of adult thinking, which include abstract and conceptual reasoning, metacognition, attention and working memory, linguistic abilities like comprehension, a tool of thought (inner speech), and a lot more.

\paragraph{Pitfalls of Commission} 

%%% RESOLVED SV: One reviewer commented negatively on the bulleted text. I think the bullets make people think the text is short and cryptic. In fact, each is a nice paragraph. I have made the decision to get rid of the bullets! Tou can reverse it if you don't like it.

The first pitfall of commission arises because researchers must use \emph{linking hypotheses} to map model performance characteristics to human performance characteristics \cite{hale-2001-probabilistic,LEVY20081126}. For example, when modeling incremental sentence processing, the log probability of the next word (given the words that come before) according to a PLM can be mapped to the time humans take to read that word \cite{cog_sci_garden_path}.
%%% SV: The next sentence includes new text you wrote. I like it so I have fully included it.
The problem is that these links are often quite distal, i.e., there is a large difference between the performance measures extracted from PLMs and those captured from humans, leaving it unclear whether PLMs are actually ``explaining'' cognitive science data. (See the next section for further discussion.)

Second, PLMs are opaque and have limited interpretability. When models and humans fail to align in their performance, the reason(s) why can be difficult to debug. This lack of interpretability is a barrier to treating these models as scientific theories \cite{kar2022interpretability,mcgrath2023can}. Furthermore, commercial models often have many tuning interventions like supervised fine-tuning, instruction tuning, and RLHF/ RLAIF. While research shows that instruction tuning can more closely align PLMs to human brain-imaging data \cite{aw2023instruction}, lack of transparency in the tuning methods used makes it difficult to understand \emph{why} they work when they do indeed work.
%%% SV: The next few sentences are new text you wrote. I like it so I have fully included it.
It is true that there exist techniques to look at the mechanistic workings of a model, for example, mechanistic interpretability. However, these are often insufficient for investigating LLM misalignment to human cognition due to their focus on low-level mechanisms, which are the wrong level of analysis for capturing the emergent, contextual, and symbolic aspects of human thought.
% use these findings in more easily available models.
% Different ways of explaining model \emph{appropriateness} for cognitive and developmental modeling may be inconsistent.

% \raj{talk about how some works are expanding the suite of tests and developing benchmarks but have yet to establish correlation among tasks.}
% \sv{The "Mellon" citation is wrong. It should be to: Anderson, J. R. (2009). How can the human mind occur in the physical universe?. Oxford University Press.}
% The use of PLMs for \emph{discovering} cognitive and developmental effects needs evidence of transferability of one PLM assessment to another in the same manner as human transferability. This means that a PLM showing impairment in one ability might perform at a human-like level in others.
    
Finally, PLMs are trained on aggregate human data (like Wikipedia), and thus, their behavior may not always reflect the broad range of different behaviors observed across individuals.
%%% SV: The next sentence is new text you wrote. I like it so I have fully included it.
Although some research has explored fingerprinting human personas with PLMs, i.e., using prompts to elicit individualistic human-like behaviors \cite{park2022social, aher2023using, potter2024hidden}, there are conflicting views on whether models can simulate these personas \cite{milivcka2024large} or not \cite{salewski2024context}.
    
%%% SV: On the other hand, let's omit this item.
% \item The human-PLM alignment scores may be wrongly interpreted when evaluating PLMs with different tasks. Cognitive alignment scores show the similarity of PLM outputs to human outputs and indicate that PLMs do not need explicit neural circuitry for these intelligence tests. They are unable to suggest   models as proxies for humans and require further testing before use \cite{eisape2023systematic, oota2023deep, hardy2023large, karamolegkou2023mapping, shah2024development}.
    
\paragraph{Pitfalls of Omission} 

Moving to pitfalls of omission, the first is that human brains and PLMs are architecturally different.
%%% SV: The next sentence includes new text you wrote. I like it so I have fully included it.
Referring to \citet{marr2010vision} on the three levels of analysis, the architectural difference lies primarily at the implementational level, where the human brain relies on biological neural networks composed of neurons and synapses, while PLMs operate using artificial neural networks implemented in silicon-based hardware.
Recent research attempts to map different aspects of PLMs (layers, attention heads, etc.) to different brain regions, seeking correspondence between model performance and functional neuroimaging measures. However, this work is in its infancy, and its viability (which requires a correspondence between NLP software and neural hardware) remains an open question \cite{hosseini2022artificial,kauf2023event}. 
% have either shown no mapping or weak mapping \cite{ oota2023deep, karamolegkou2023mapping, van2020brain, cog_sci_garden_path, nikolaus2024modality}.

Second, most studies evaluating the cognitive alignment of PLMs focus on a narrow range of cognitive abilities and overlook correlations with other abilities. This is in line with the \emph{experimental} approach to human behavior -- but at odds with the \emph{differential} or \emph{psychometric} approach, which has also proven to be important \cite{Cronbach1957TheTD}. While recent research is developing larger benchmarks and expanding the evaluation of PLMs to suites of tests \cite{coda2024cogbench,chang2024benchmarking}, there has been almost no attention paid to the correlations between the various tests. By contrast, psychometric approaches to human intelligence put the focus on the correlations across tests of a broad range of cognitive abilities: mathematical, verbal, spatial, fluid, and so on \cite{snow1984topography, schneider2012cattell}.
This differential view is also in contrast with unified theories of cognition, the precursors to Artificial General Intelligence, that attempt to model all cognitive abilities within a single computational framework \cite{anderson2009can, varma2011criteria}. 
    
\paragraph{Challenges of Development} 

Finally, we consider the pitfalls that come with treating PLMs as developmental science theories, i.e., of the progressions in children's thinking over time.

First, PLM checkpoints are snapshots or fingerprints of the data on which they are trained. Most research only looks at final model checkpoints and evaluates cognitive alignment to adult thinking. Equally important is the question of whether, as language models observe more and more data, their performance is aligned to that of older and older children \cite{warstadt2024artificial,frank2023bridging, shah2024development}. Developmental alignment is frequently overlooked because of the general unavailability of intermediate training checkpoints or because of resource constraints. This limits our understanding of the developmental fidelity of model training. However, recent open-source language modeling efforts to make available these checkpoints represent a promising opportunity to study developmental progressions \cite{biderman2023pythia,liu2023llm360,OLMo}. 

Second, there are large differences in the nature of the data observed by PLMs versus those experienced by humans. PLMs are trained on magnitudes more textual data than the number of words seen by children \cite{huebner-etal-2021-babyberta,hosseini2022artificial,conll-2023-babylm,bhardwaj2024pretraining}. That said, children learn from input from multiple senses \cite{smith2005development}, whereas models are not embodied in nature \cite{chemero2023llms,cuskley2024limitations,birhane2024large}. Here, the emergence of vision-language models offers the potential to bridge the gap between the disembodied nature of PLMs and the multisensory learning of humans. However, this approach still faces certain limitations. For example, research suggests that while models can process simple visual features like color and size, models struggle with complex spatial and numerical reasoning tasks when applied to novel contexts or objects \cite{yiu2024kiva}.
        
Finally, for studies that have evaluated the developmental alignment of PLMs, the developmental trajectories observed in models might be artifacts of the pre-training order \cite{shah2024development}.
%%% SV: The final sentence is a bit complex. If you can figure out how to neatly divide it into two sentences that are each more manageable, that would be great.
Studies are needed to analyze training corpora and assess the impact of training curriculum design. This includes examining the sequence and nature of data presented during pre-training to distinguish genuine developmental progressions from artifacts introduced by training strategies.

\begin{tcolorbox}[keytakeaways,title=Pitfalls]
   Using PLMs as cognitive models potentially brings pitfalls of commission and omission and requires attention to the challenges of development. This means that researchers should be cautious when assessing the suitability of these models as proxies for human cognition and its development.
\end{tcolorbox}

\section{Linking Hypotheses - Mapping Model Performance to Human Performance}
\label{linking}

The “outputs” of PLMs are often quite different from the behavioral measures that cognitive scientists collect in their experiments and evaluate their theories against. Researchers use various linking hypotheses to map PLM performance to human performance measures. These assumptions necessarily define -- and potentially limit -- the strength and validity of the alignment.
It is important to critically evaluate these linking hypotheses because they structure how we interpret the models' cognitive capabilities. Below, we review different approaches to mapping indices of model behavior to human performance.
% The most common approach is mapping human response times to model uncertainty.
% While all of these linking hypotheses are simple in theory, they have multiple possible operationalizations. 
% \sv{The prior sentence is pretty vague at the point and can be deleted here without problem.} 

% \raj{Latent Representation:}
\paragraph{Similarity computations}

Many cognitive tasks require people to judge the similarity of two items. Examples range from Shepard’s classic psychophysical studies of people’s similarity judgments of perceptual stimuli (e.g., circles of differing diameters) \cite{shepard1978cognitive} to Griffiths' current Bayesian and neural network models of how people judge the similarity of an exemplar to a category prototype \cite{l2008bayesian}.

Human similarity judgments can be directly modeled by computing the similarity between the corresponding representations in a PLM's latent space. 
This can be via cosine similarity or other metrics \cite{turney2005corpus}.
%%% RESOLVED SV: For the sentence below, I would see Sid's paper for better cog sci citations, e.g., to Rosch. RAJ: Removed one of the two bhatia and ritchie citaitons
One example of similarity computations in cognitive tasks comes from modeling the typicality effect, which is the finding that people regard some members as ``better’’ examples of a category than others \cite{rosch75, bhatia2022transformer}.  The typicality of an exemplar is commonly defined as the proportion of humans that produce it when asked to enumerate the exemplars of a category, with larger proportions indicating greater typicality. 
% PLMs proxy this ranking by capturing the uncertainty of an item with different operationalization.
In language models, the typicality of an exemplar for a category can be estimated by encoding the exemplar name as a string, passing it through the language model, and obtaining the corresponding word embedding. Thereafter, the similarity between this exemplar vector and the category prototype (e.g., the embedding obtained by encoding the category label as a string and passing it through the model) is calculated, with a higher value indicating that the exemplar is more typical \cite{misra2021language,cog_sci_typicallity}. 
% Similarly, works have modeled various ranking-based tasks to PLM uncertainty \cite{misra2021language,cog_sci_typicallity}.  

Other cognitive tasks require comparing or discriminating between two items. There, a common linking hypothesis is that the greater the similarity between the items in the model's latent space, the longer the comparison/discrimination time.
Coming back to the example of a latent mental number line, \citet{shah2023numeric} map the time it takes to compare which of two numbers is greater to the similarity of their PLM encodings. The linking hypothesis is that the greater the similarity of (i.e., the less discriminable) the two encodings, the longer the predicted time for humans to judge which one is greater. Under this mapping, PLMs show the `distance, size, and ratio effects observed in humans \cite{moyerTimeRequiredJudgements1967} and thought to be behavioral signatures of possessing a ``mental number line''.

More broadly, research shows the neural plausibility of vector-based representations in capturing human conceptual properties, including compositionality, feature computation, relational reasoning, and symbolic processing \cite{piantadosi2024concepts}.

There are two common obstacles to using similarity as a linking hypothesis. The first is that doing so requires models that make available the latent representations for similarity computations. 
% \newtext{Often, the latent representations are unavailable for commercial models \cite{openai2023gpt4}}.
The second is that
% The last approach, the closeness judgment problem formulation, looks at latent space. This approach enables us to view the internal representation of the model, compare it to other methods like word2vec, and visualize PLM semantic spaces. Yet, 
this method suffers from problems due to tokenization. Humans use different granular units (words or subwords) compared to PLMs, which use tokens. This introduces misalignment -- the mapping is inconsistent as one unit of text for humans may be mapped to two units for PLMs. For example, the word ``Nine'' for humans is represented as \emph{N, \#ine} in some PLMs such as Pythia \cite{biderman2023pythia}. 
%%% SV: I don't think we need the final sentence so I commented it out.
% This makes them harder to operationalize and interpret.
% \raj{Isn't surprisal a psych term?}

\paragraph{Surprisal values}

One way of quantifying the uncertainty of model generations is in terms of the summation of their surprisas, i.e., their negative log probabilities.
% For example, for the typicality experiments, \citet{shah2024development} compare the probability of the generated sequence in PLMs, ``an {item} is a {category}’’, as a proxy for surprise.
A common linking hypothesis is that higher surprisal values correspond to longer human response times. For instance, studies of reading \cite{rambelli2024neural,ivanova2024elements} and categorization \cite{misra2021language} have found evidence for this correspondence. Similarly, \citet{shain2024word} use PLMs to demonstrate strong surprisal predictability estimates of human reading times. Research also shows that surprisal values provide a better match to human plausibility judgments than prompts \cite{ivanovalog}. 
Finally, relative surprisal has been used to distinguish grammatical and ungrammatical sentences \cite{warstadt2020blimp}. In this case, surprisal enables direct comparison of the right answer with all possible candidate answers in a deterministic manner because PLMs will generate sequential probabilities for all strings, whether they are grammatical or not. 
% \sv{The end of the sentence above is my attempt to rewrite the following fragment, which I found difficult to understand, in part because of the double negative: "i.e., there is no chance that the PLM will not calculate the sequential probability for a candidate string."} \raj{got it, also add in  the raven's, SAT or some IRT workk}

A problem with surprisal-based approaches is that they show high context sensitivity, i.e., the alignment of PLMs to humans often depends on the framing and the structure of the original prompt.
%%% RESOLVED - SV: I got lost on how the next 2 sentences illustrate the claim of the first sentence of the paragraph. RAJ: Added new text
For example, in mathematical tasks, models tend to achieve better alignment with human reasoning when they are prompted to generate longer texts and step-by-step reasoning chains \cite{jiang2024llms}. This sensitivity to prompt structure makes it challenging to operationalize variations in PLM outputs in a way that mirrors the natural variances observed in human responses to stimuli.
% In the case of many mathematical tasks, forcing a model to generate longer texts and reasoning chains leads to better human alignment \cite{jiang2024llms}. Along similar lines, it is difficult to operationalize variances in PLM outputs akin to variances in human responses to stimuli. 
In some cases, surprisal is a weak proxy for human behavior: \citet{van2021single} find that estimates of the time cost of word predictability derived using surprisal methods underestimate the magnitude of human garden path effects during processing of temporarily ambiguous sentences. More generally, surprisal fails to explain syntactic disambiguation difficulty \cite{huang2024large}. These findings highlight some failure modes of surprisal as a linking hypothesis and reveal the need to establish the empirical plausibility -- goodness of fit score -- of each linking hypothesis on each task.
%%% SI I like the last sentence, which is new, and accepted it. However, I am now worried that the final 3 sentences of this paragraph are summary in nature, which is too many. It feels redundant.
Surprisal makes assumptions about the relevance of probability distribution shifts to human sentence parsing and the relevance of information-theoretic measures to human cognition, which may be wrong or non-neutral and might be partially due to how surprisal is framed.

% \sv{I don't understand this sentence. Perhaps this ie because "robustness" is a vague words and I can't retrieve the sense in which it is intended here. Can you flesh out this criticism a bit?}

% Here are popular ways of operationalizing such assumptions:
%     \begin{itemize}[leftmargin=*]
%         \item Closeness judgment problem: For the problem involving comparisons, uncertainty is modeled as the similarity between the two items in the latent space. Higher similarity implies more time to distinguish/ make the judgment. In this method, researchers calculate the cosine similarity (or other similarity metrics) between the obtained latent representations for item 1 and item 2 (items 1 and 2 may be heterogeneous). This requires models where the latent representations are readily available.
% 
%         \item Surprisal values: Another way of quantifying the uncertainty of model generations is in terms of the summation of logarithmic probabilities. For example, for the typicality experiments, \citet{shah2024development} compare the probability of the generated sequence in PLMs, ``an {item} is a {category}’’, as a proxy for surprise.
%         
%         \item Prompting: This method quantifies surprise differently. PLM generation is probabilistic and the same model can give different results across inference runs. A PLM is prompted to follow the same exercise as a human multiple times and generate a probability distribution over the output space. The probability of the correct output is then mapped to model confidence.
%     \end{itemize}

\paragraph{Prompting}

PLM generation is probabilistic and thus, non-deterministic, i.e., the same model can give different results across inference runs. In prompting, PLM can be instructed to follow the same exercise as a human multiple times to generate a probability distribution over the output space. The probability of the correct output is then mapped to model confidence. Thus, prompting is very ``proximal'', enabling direct comparison of the generated behavior of PLMs with that of humans and reducing (or even eliminating) the need for relatively ``distal'' and indirect linking hypotheses \cite{patel2021mapping, webb2023emergent, zahraei2024wsc+}.
% Prompting allows for more \emph{flexibility in the input choice}, furthermore, the input can be presented to the PLM in the same manner as humans.
For example, to evaluate a model on the BLiMP benchmark \cite{warstadt2020blimp}, a PLM can be prompted with:

\noindent \textcolor{darkgray}{Prompt: Which statement is grammatically correct? Your response must be "1" or "2".}

\noindent \textcolor{blue}{1. Noah likes to swim.}  {} {}  {} {}  {} {}  {} {}  {} {}  {} {} \textcolor{red}{2. Noah likes to.}

\noindent The PLM can directly generate ``1'' for the correct answer, and its probability of doing so across inference runs can be directly compared with the probability of humans making the correct choice. This is much closer to the task humans perform than the more typical method of comparing the relative surprisal of the two sentences.  

Another benefit of prompting is that it allows for \emph{variable output length}. Tasks that benefit from this flexibility, like commonsense reasoning, are more suited for prompting \cite{yasunaga2023large}. This flexibility allows us to directly ask the model about its current state. For example, to measure the incremental semantic understanding of temporarily ambiguous sentences, \citet{cog_sci_garden_path} present a dichotomous verification sentence to the model after each word. The shifting probabilities of the sentences consistent with different semantic interpretations directly evidence the model's online handling of ambiguity and response to disambiguating information. 
PLM prompting enables direct probing, which is often not possible in humans with behavioral measures (i.e., it would be disruptive to ask a human reader after each word to verify sentences), or even with using neuroimaging measures, which suffer their limitations to ``look under the hood''.

Another example is the SAT analogical reasoning tasks \cite{turney-2013-distributional}. These are of the form A:B::? For example, consider the following SAT analogical reasoning problem:

\noindent \textbf{Analogy:} Runner : Marathon :: ?\\
\textbf{Options:}
\begin{multicols}{2}
\begin{itemize}[leftmargin=*]
    \item Envoy : Embassy 
    \item Martyr : Massacre 
    \item Oarsman : Regatta 
    \item Horse : Stable
\end{itemize}
\end{multicols}

\noindent Such multiple choice question-based choice tasks can be operationalized using similarity, surprisal, or prompt-based reasoning. In the prompting case, we can force the generations to adhere to the goal: \textit{answer the correct analogy in this output format: \{"A":"B"::"C":"D"\}.}

%%% SV: Give an example of a Turney analogy problem.
% Runner: Marathon ::
% \begin{table}[!h]
%     \centering
%     \caption{Example SAT analogical reasoning problem.}
%     \label{tab:sat}
%     \resizebox{0.3\textwidth}{!}{%
% \begin{tabular}{ m{1.3cm}  m{4cm} }
%     \hline
%     \textbf{Analogy:} & Runner : Marathon :: ? \\  \hline

%        & (A) envoy : embassy \\
%         & (B) martyr : massacre \\
%       \textbf{Options:} &   (C) oarsman : regatta \\
%         & (D) referee : tournament \\
%         & (E) horse: stable
% \\
%     \hline 
% \end{tabular}
% }
% \end{table}

Although it is quite ``natural'', there are several limitations to the prompting approach. First, similarly to the surprisal method, PLMs are \emph{not robust to the prompt format.}  Answering prompts requires PLMs to (1) understand the prompt and (2) know the answer. 
%%% RESOLVED SV: The Guo citation next is not resolving properly!
Often, small variations in the input prompts lead to large changes in human alignment \cite{guo2024understanding}, and sometimes maximum performance characteristics are obtained on gibberish prompts \cite{deng2022rlprompt}. Second, PLMs may output an answer beyond the given candidate options for the prompting approach. For example, the outputted answer to the grammaticality judgment prompt above could be "It would be irresponsible to imply that the grammatical structure of a sentence is inconsequential, as clear communication is fundamental for safety and understanding" -- output of an existing PLM observed by \citet{cai2024antagonistic}. No human would make this response. In the same vein of errors, when asked to follow complex human-like experimental procedures, prompting methods sometimes fail in maintaining output format consistency, making it hard to use this method. Third, research also points towards certain tasks where prompting lacks efficacy and argues against the exact metalinguistic prompting approach presented above \cite{hu2023context,hu2024auxiliary}. 

%%% SV: I eliminated the italicization of this paragraph. I also accepted the last sentence, which is new text you wrote.
Overall, we see that some linking hypotheses are better suited for certain tasks than others, and sometimes, specific linking hypotheses fail to show any human behavior profile. Furthermore, it is hard to find the direction of error when the linking hypothesis fails to show a human behavioral profile, i.e., it is challenging to understand misalignment - whether it is due to over-simplified or over-complicated assumptions.

\begin{tcolorbox}[keytakeaways,title=Linking Hypothesis]
   Our findings 
   % highlight the strengths and failure modes of different linking hypotheses and 
   reveal the need to establish the strong empirical plausibility -- goodness of fit score -- for each linking hypothesis on each task.
\end{tcolorbox}

\section{Criteria for Evaluating and Developing PLMs as Scientific Models}

PLMs are increasingly being evaluated as models of cognitive and developmental science phenomena. In light of the discussion above, the review of some of the popular assumptions (linking hypotheses), and the common pitfalls outlined, we propose two sets of criteria for using PLMs for this purpose.
While some may seem obvious or trivial, we believe that it is important to explicitly document them, given the evolving complexity and scope of PLMs. Explicit criteria provide a basis for interpreting results meaningfully, identifying limitations, and ensuring methodological rigor in their application as cognitive models.

\subsection{Appropriateness}

The first set concerns the \emph{appropriateness} of PLMs as scientific tools for modeling cognitive and developmental phenomena:

%%% SV: One reviewer commented negatively on the bulleted text. I think the bullets make people think the text is short and cryptic. In fact, each is a nice paragraph. I have made the decision to get rid of the bullets! Tou can reverse it if you don't like it.

\paragraph{Design multiple experiments to test alignment to each cognitive or developmental phenomenon}

PLMs may track human performance characteristics well under one linking hypothesis on one type of test. However, this alignment may just be an artifact, for example, of pre-training data contamination, i.e., the accidental inclusion of evaluation samples in the pre-training data.
%%% SV: I accepted the new text you wrote, which comes next.
We recommend \emph{empirical triangulation} -- conducting more experiments evaluating the same cognitive/developmental phenomena -- to establish stronger empirical plausibility.

\paragraph{Use multiple methods to interpret PLM successes and failures}

PLMs lack explainability and interpretability due to their large size \cite{mcgrath2023can}. 
% \sv{The year for the McGrath reference is not rendering.}
Some methods for PLM interpretation are often better than others. For example, in the experiments conducted by \citet{cog_sci_garden_path}, incrementally constructed parse trees provided a better account of PLM alignment than the information from attention weights. 
% \sv{The prior sentence is where I suggested moving some or all of this sentence from section 4: "In addition to the direct probing, one can directly recover the implicit parse tree after each word to measure incremental syntactic understanding \cite{manning2020emergent}."} 
It is typically not clear in advance which evaluation metric will be most insightful for a given set of models and cognitive tasks, \emph{necessitating a multi-pronged approach.}

\paragraph{Test the path-dependency of PLMs for developmental alignment}

The claim that the final model state of a PLM approximates adult performance leads to the question of the path by which it arrived there. Ideally, the model's performance improvements over training should also track the progression of cognitive abilities over development \cite{elman1996rethinking, bengio2009curriculum}. This would support researchers exploring the scaling of training data and model size in their investigations of human development.
%%% SV: I accepted the new text you wrote, which comes next.
Inspired by \cite{tan2024devbench}, we encourage research into building age-aligned developmental benchmarks by sampling data aligned with children's learning trajectories.
    
    %%% SV: Copy the text of the last sentence earlier, where I asked you to provide more detail in the Li et al. (2024) paper.
    
    %%% Raj: On direct probing sometimes being more useful than other methods?
    
    %%% SV: This concern is not a big one for cognitive scientists so I commented it out to save space. However, if you believe the EMNLP audience will definitely want it stated explicitly, then add it back in.
    %%% SV: Note that we make this point in the Ethical Considerations section, giving a second reason to cut it here.
    % \item \textbf{Remember that PLM alignment does not indicate human replacement:} Any potential discovery using PLMs about human cognition or development needs to be tested with humans for validation.

%%% SV: A bit abstract. Can you give an example to make especially the second sentence more specific.
\paragraph{Control for tuning techniques}

PLMs are often tuned on specific data and in different ways, such as Instruction Tuning, Reinforcement Learning from Human Feedback (RLHF), etc. The technique used is not incidental. Rather, it can influence model behavior, which is important if the model's output centers around tuning goals rather than developing a representation of world knowledge. Empirical evidence suggests that tuning methods result in better aligning PLMs with human cognition \cite{aw2023instruction}.
%%% SV: I accepted the new text you wrote, which comes next.
We suggest evaluating cognitive alignment across multiple PLMs tuned with different objectives (e.g., instruction-tuned vs. RLHF-tuned).
% \sv{You "tuned" this text and I find it much more readable. I made some edits to make it flow a bit better. Please correct any errors I may have introduced.}

%%% SV: Try to connect the language of the next item to the language we developed in the Linking Hypotheses section above. Right now, it seems quite different.
\paragraph{Remember the linking hypothesis}

Adapting human experimental materials to textual counterparts that match the required modalities for PLMs requires making certain assumptions (refer to different operationalizations in section \ref{linking}). These assumptions are not neutral, but rather part of why models may or may not align with human performance. For this reason, they need to be well-documented and explored in their own right. Additionally, transparency in selecting such assumptions helps establish the credibility and replicability of findings.

\paragraph{Establish task correlations}

%%% RESOLVED SV: I made a pretty big change to the end of this paragraph, commenting out the sentences I replaced.
Inspired by \citet{snow1984topography}, who study the correlations in human performance across different psychometric tests of intelligence, we propose a similar cross-task PLM evaluation paradigm. The goal will be to evaluate whether the pattern of cross-task correlations observed in humans is also produced by PLMs.
% This paradigm will allow us to make estimates of model performance alignment with human behaviors on a set of novel tasks given PLM testing profiles on other tasks. This approach enables us to understand the \emph{broader capabilities and generalization} of the model across different types of cognitive challenges. 

\paragraph{Embodiment and interactiveness}

Many researchers have advocated for the addition of more modalities in the pre-training process, such as vision, touch, etc., to emulate the learning environment of the child \cite{cuskley2024limitations}. While purely theoretical and speculative, the increasing ``embodiment'' of pre-training and the addition of external interactions may yield more substantial human alignment.

\subsection{Development of PLMs for Cognitive Modeling}
% \raj{Discuss this section}
The second set of criteria is for guiding the \emph{development} of PLMs as credible accounts of cognition and its development. This is a more open-ended task, and the following can be considered as mere suggestions to researchers:

%%% SV: I moved this point up here and commented out the one that was here, which makes the same point. Feel free to merge the text if you like some of what has now been removed
\paragraph{Pre-training data may benefit from developmentally plausible corpora} 

PLMs should be evaluated at regular intervals of pre-training to assess their potential developmental alignment. This step is often overlooked in current studies of cognitive alignment. This includes training on a curriculum based on the known developmental trajectories of knowledge and skill acquisition \cite{bhardwaj2024pretraining, warstadt2024artificial, frank2023bridging,hu2024findings}. For example, corpora could be ordered or sampled based on the age-of-acquisition of the words their text contains \cite{huebner-etal-2021-babyberta, Portelance2023PredictingAO}. Informed pre-training will allow us to better understand the developmental alignment of models.

% \paragraph{Evaluation techniques should follow the appropriateness criteria above}
%  Evaluation should also focus on constructing developmental benchmarks, such as the one presented by \citet{yiu2024kiva}, where behavioral profiles of 3-year-old children are captured and mapped to visual language model outputs.

% \item \textbf{Development should not always focus solely on topline model performance.} Children's cognitive intelligence develops in stages, with progression in skills like reasoning, memory, problem-solving, and language. Analysing the peak model performance as a function of training can overlook the step-by-step incremental process of child learning. \citet{tan2024devbench} look at divergence between model predictions and human performance, as an alternative to pure topline performance
    
\paragraph{PLMs can first be tuned on a small number of ``core'' cognitive tasks and then evaluated on a broader range of cognitive tasks}

For example, typicality experiments \cite{cog_sci_typicallity, misra2021language} could be used to preference-tune PLMs using reinforcement learning techniques. This might result in better cognitive alignment to a large number of tasks that all rely on human-like semantic representations \cite{vazquez-martinez-2021-acceptability, javier-vazquez-martinez-etal-2023-evaluating}.
%%% SV: I accepted the new text you wrote, which comes next. However, see my comment earlier that maybe this is too much, and should be commented out.
This builds off the promise of \emph{generalization and transferability} for model alignment, where alignment on $n$ tasks leads to predictive utility and alignment on $n+1\text{th}$ task \cite{binz2024centaur}.

% \raj{Practical implications: Evaluation standards for models, methodological and technical limitations}
\section{Conclusion}

This paper advocates for the use of Pre-trained Language Models (PLMs) as theoretical tools for investigating human cognition and its development through the three-stage model shown in Figure \ref{fig:sufficieency}. In this advocacy, we are not alone \cite{mcgrath2023can, frank2023bridging, warstadt2024artificial,mahowald2024dissociating}. However, at the same time, we caution researchers toward the informed use of PLMs when making theoretical claims in the cognitive and developmental sciences. We have highlighted common pitfalls in this enterprise, reviewed the different assumptions (i.e., linking hypotheses) used by researchers to map PLM performance to human performance, and outlined criteria for evaluating and developing PLMs as credible models of cognition and cognitive development.
%%% SV: I accepted the new text you wrote, which comes next.
These criteria are intended to guide researchers in designing robust experiments (particularly using the three-stage criteria as a framework for experiment design), interpreting PLM behaviors accurately, and increasing fidelity to human data. Given the constantly evolving nature of the field, we call for researchers to continuously refine and expand these guidelines to match new advancements in NLP and cognitive science.
% Lastly, the paper outlines the \emph{criteria} for using and developing PLMs to be credible accounts of psychological sciences. 

% This paper advocates for the use of Pre-trained Language Models (PLMs) as theoretical frameworks for understanding cognitive intelligence and its development. While recognizing the potential of PLMs to offer valuable insights, the paper urges researchers to exercise caution and be well-informed about the inherent limitations and challenges. By highlighting common pitfalls and critically examining the assumptions underlying the mapping of human and PLM performance measures, we provide a nuanced perspective on the use of PLMs in psychological sciences.

\section{Limitations}
 
(1) The paper highlights common pitfalls, linking hypotheses, and evaluative criteria while using PLMs for cognitive modeling. These constitute a set of sound views to aid new researchers in the field. They do not exhaustively cover every pitfall, hypothesis, or criterion.
(2) The suggestions in this work are good-to-have practices that support using PLMs for open cognitive and developmental science. No one-answer-fits-all approach is possible. NLP is a developing field, and we recommend articulating newer guidelines and practices as newer and ever-larger PLMs are trained and deployed.
(3) Our work calls for language technologies for the psychological sciences and provides criteria for developing credible accounts of cognition and cognitive development.
%%% SV: I accepted the new text you wrote, which comes next.
Despite providing general guidelines, our work is theoretical and does not conduct experiments or offer empirical evidence of performance comparisons or other quantitative measures. In section 5, we suggest criteria by building upon and refining prior works based on empirical observations from the field.
(4) PLMs, as described in this paper, are artificially designed systems that hope to reveal mechanisms of naturally evolving ones. Thus, researchers should keep the final goal in mind and not mistake a technological tool for an absolute source of insight.
(5) This paper focuses on using PLMs for behaviorally benchmarking cognitive profiles of neurotypical and developmentally typical individuals. PLMs should also be used for simulating the behaviors of impaired individuals to better support them.
%%% SV: I accepted the new text you wrote, which comes next.
(6) We do not address broader ethical implications of modeling cognitive impairments, limited cultural generalizability of PLMs trained on Western-centric corpora, and other factors outside the scope of our arguments.

\section{Ethical Considerations}

There are no significant risks associated with conducting this research beyond those associated with working with PLMs. There may be risks in misinterpreting the criteria enlisted in this study. The suggestions in this study are one-way: we wish to find human performance characteristics and behaviors in PLMs to help model psychological sciences and, in the future, to aid people with cognitive impairments. We do not advocate for developing PLMs to replace humans or suggest ways to reach Artificial General Intelligence. PLMs are experimental technologies, and future work using these models should be conducted cautiously.

\bibliography{custom}

% \appendix
% \section{List of comprehensive list of domains where PLMs have been used as Cognitive Science Models}
% \input{latex/papers_table}
% % \label{sec:appendix}

% This is an appendix.

\end{document}